\documentclass{article} % For LaTeX2e
\usepackage{iclr2025_conference,times}

\usepackage{hyperref}
\usepackage{url}
\usepackage{graphicx}
\usepackage{amsmath}
\usepackage{mathrsfs}
\usepackage{wrapfig}
\usepackage{booktabs}       % professional-quality tables
\usepackage{amsfonts}       % blackboard math symbols
\usepackage{nicefrac}       % compact symbols for 1/2, etc.
\usepackage{microtype}      % microtypography        % colors
\usepackage{multirow}
\usepackage{multicol}
\usepackage{makecell}
\usepackage{mathtools}
\usepackage{xspace}
\usepackage{amssymb}
\usepackage{algorithm}
\usepackage{algorithmic}
\usepackage{colortbl}
\usepackage{arydshln}
\usepackage{float}
\usepackage{subfigure}
\usepackage{xcolor}
\usepackage{lineno}
\usepackage{enumitem}
\definecolor{citecolor}{HTML}{0071BC}
\definecolor{linkcolor}{HTML}{ED1C24}
\definecolor{grey}{HTML}{999999}
\definecolor{green}{HTML}{ABD1BC}
\definecolor{lightblue}{HTML}{B0C4DE}
\definecolor{purple}{HTML}{E3BBED}
\definecolor{orange}{HTML}{ffdab9}

\definecolor{lightmauve}{rgb}{0.86, 0.82, 1.0}

\hypersetup{
colorlinks=true,
linkcolor=linkcolor,
citecolor=citecolor,
}

\title{\makebox[0pt][l]{\hspace{-0.2em}\raisebox{-0.2em}{\includegraphics[height=1.5em]{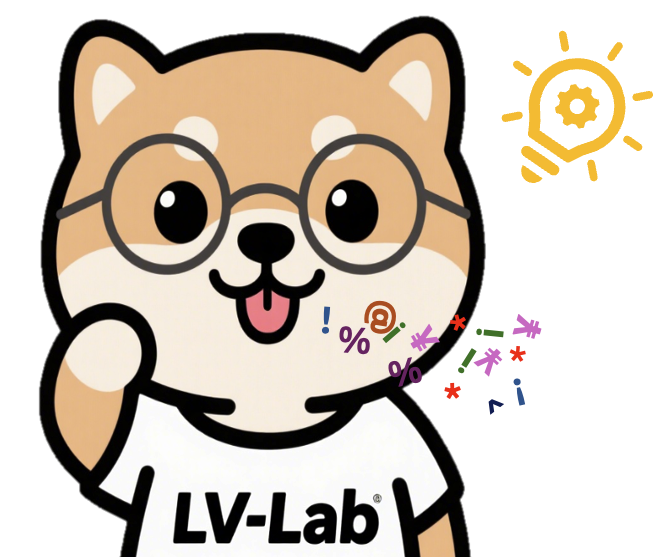}}}%
\hspace{1.4em}MINI-OMNI-REASONER: Token-Level \\ Thinking-in-Speaking in Large Speech Models}

\author{
Zhifei Xie\thanks{Equal contribution \quad $^\dagger$Corresponding authors } \quad 
Ziyang Ma$^{*}$ \quad Zihang Liu \quad Kaiyu Pang \quad Hongyu Li \quad Jialin Zhang \\  
\textbf{ Yue Liao}$^{\dagger}$ \quad \textbf{Deheng Ye}$^{\dagger}$ \quad
\textbf{Chunyan Miao}$^{\dagger}$ \quad \textbf{Shuicheng Yan}$^{\dagger}$ 
\\
Nanyang Technological University \quad National University of Singapore \quad Tencent \\
\href{mailto:zhifei001@e.ntu.edu.sg}{\textcolor{black}{\texttt{\{Zhifei001, ziyang012\}@e.ntu.edu.sg}}} \quad
\href{mailto:liaoyue.ai@gmail.com}{\textcolor{black}{\texttt{liaoyue.ai@gmail.com}}} \\
\href{mailto:dericye@tencent.com}{\textcolor{black}{\texttt{dericye@tencent.com}}}  \quad
\href{mailto:ascymiao@ntu.edu.sg}{\textcolor{black}{\texttt{ascymiao@ntu.edu.sg}}} \quad
\href{mailto:yansc@nus.edu.sg}{\textcolor{black}{\texttt{yansc@nus.edu.sg}}}  \quad
\\ \\ 
\normalsize ~\url{https://github.com/xzf-thu/Mini-Omni-Reasoner}
}
\iclrfinalcopy % Uncomment for camera-ready version, but NOT for submission.
\begin{document}

\maketitle

\begin{abstract}
Reasoning is essential for effective communication and decision-making. While recent advances in large language models (LLMs) and mul
timodal models (MLLMs) have shown that incorporating explicit reasoning significantly improves understanding and generalization, reasoning in large speech models (LSMs) remains in a nascent stage. Early efforts attempt to transfer the “thinking-before-speaking” paradigm from textual models to speech. However, this sequential formulation introduces notable latency, as spoken responses are delayed until reasoning is fully completed, impairing real-time interaction and communication efficiency. To address this, we propose Mini-Omni-Reasoner, a framework that enables reasoning within speech via a novel “thinking-in-speaking” formulation. Rather than completing reasoning before producing any verbal output, Mini-Omni-Reasoner interleaves silent reasoning tokens with spoken response tokens at the token level. This design allows continuous speech generation while embedding structured internal reasoning, leveraging the model’s high-frequency token processing capability. Although interleaved, local semantic alignment is enforced to ensure that each response token is informed by its preceding reasoning. To support this framework, we introduce \textsc{Spoken-Math-Problems-3M}, a large-scale dataset tailored for interleaved reasoning and response. The dataset ensures that verbal tokens consistently follow relevant reasoning content, enabling accurate and efficient learning of speech-coupled reasoning. Built on a hierarchical Thinker–Talker architecture, Mini-Omni-Reasoner delivers fluent yet logically grounded spoken responses, maintaining both naturalness and precision. On the Spoken-MQA benchmark, it achieves a +19.1\% gain in arithmetic reasoning and +6.4\% in contextual understanding, with shorter outputs and zero decoding latency. These results demonstrate that high-quality reasoning and real-time spoken interaction can be effectively unified in a single framework. 
\end{abstract}

\begin{figure}[t]
    \centering
    \includegraphics[width=1\linewidth]{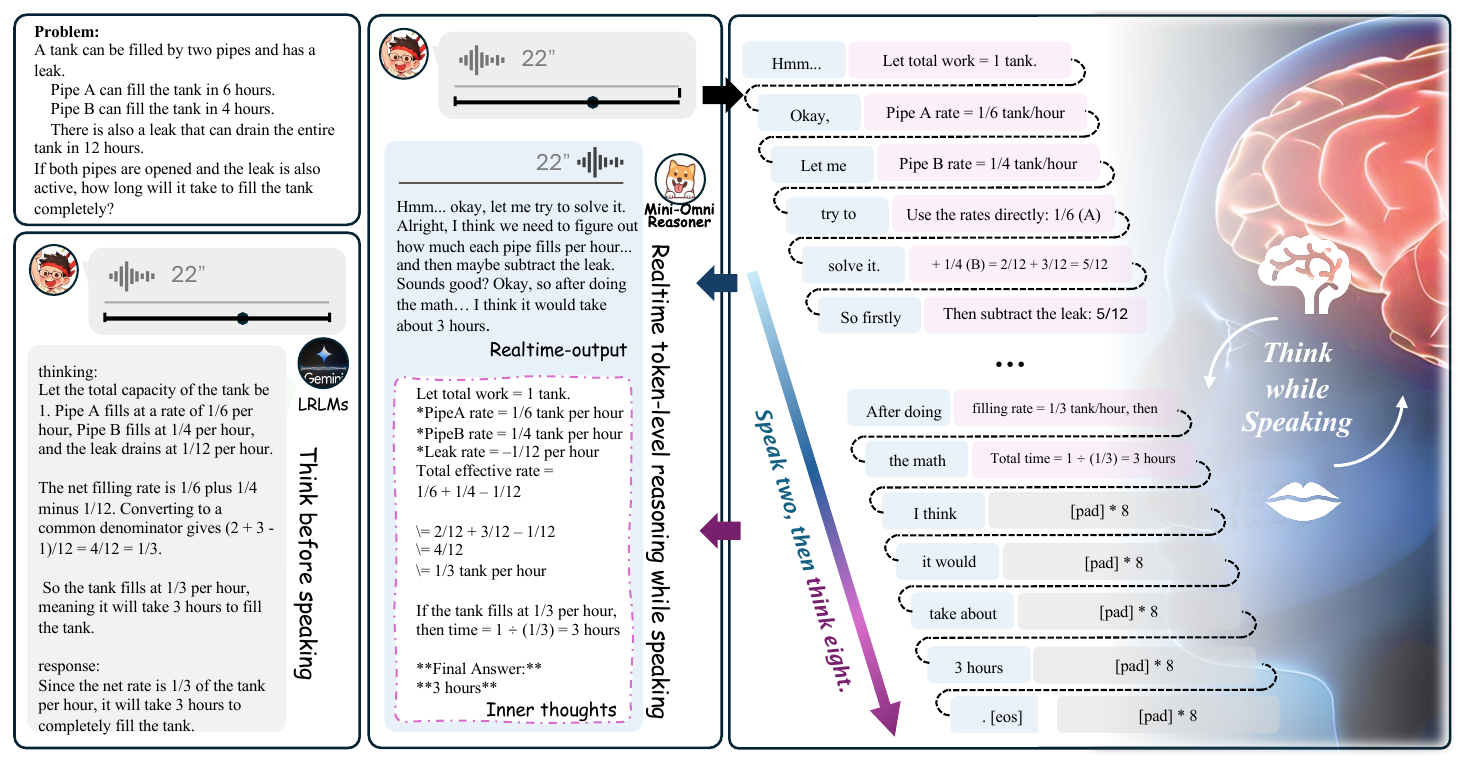}
    \vspace{-3mm}
    \caption{Comparison between the traditional \emph{“thinking-before-speaking”} reasoning paradigm and our proposed \emph{“thinking-in-speaking”} paradigm.
The traditional paradigm requires completing the entire reasoning process before producing any spoken output, resulting in long latency or forcing the model to speak out verbose reasoning before delivering the actual answer.
In contrast, \emph{“thinking-in-speaking”} paradigm interleaves high-frequency internal reasoning with continuous speech generation, enabling the model to deliver timely and informative responses while maintaining reasoning quality. This design leverages the mismatch between model-side inference throughput and audio playback constraints to reduce latency and improve listener experience without sacrificing inference depth.
    }
    \label{fig:fig1}
\end{figure}
\section{Introduction}
Reasoning is a fundamental faculty of human cognition, enabling precise, logically structured, and contextually grounded understanding of the external world~\citep{reasoning}. In natural communication and decision-making, humans frequently engage in internal deliberation prior to verbal expression, a strategy shown to enhance the factual accuracy, completeness, and reliability of responses. 
Inspired by this cognitive mechanism, recent advances in large language models (LLMs)~\citep{o1,skywork-r1,qwq,deepseek-r1} have formalized this strategy into the computational paradigm of \emph{“thinking-before-speaking”}. In this formulation, models are prompted to construct an explicit and logically structured reasoning trace, which subsequently informs the final response. This reasoning-first formulation has demonstrated substantial benefits across a range of language tasks that demand structured explanation and logical consistency, such as arithmetic reasoning and compositional question answering.

While the \emph{“thinking-before-speaking”} paradigm has proven effective in textual domains, its direct extension to speech interfaces encounters inherent modality-specific constraints.  Text affords \emph{spatially parallel information access}: readers can scan, skip, and selectively attend to different portions of content, enabling efficient comprehension of extended reasoning sequences at high reading speeds. In contrast, speech is consumed sequentially over time, constrained by the fixed-rate, streaming nature of auditory perception and human cognitive processing.  Speaking out the full reasoning trace before delivering an answer may burden listeners with verbose or low-utility content, delaying access to the core response. Conversely, keeping the reasoning process silent leads to significant initial latency, as the model must complete its internal reasoning before producing any spoken response, thereby compromising interaction quality.

To bridge the gap between language reasoning and speech communication, where the conventional \emph{“thinking-before-speaking”} paradigm proves ineffective for real-time spoken interaction, we introduce \textsc{Mini-Omni-Reasoner}, a novel speech reasoning framework founded on the principle of \emph{“thinking-in-speaking”}. As shown in Figure~\ref{fig:fig1}, this formulation enables large speech-language models (LSLMs) to perform high-frequency internal reasoning in tandem with the real-time generation of semantically informative spoken tokens. By decoupling the temporal resolution of internal inference from that of speech emission, our framework supports low-latency, cognitively aligned spoken interaction without sacrificing the depth, rigor, or interpretability of the underlying reasoning process.

\textsc{Mini-Omni-Reasoner} instantiates the \emph{“thinking-in-speaking”} paradigm through an interleaved generation scheme that capitalizes on the discrepancy between model-side inference throughput and real-time audio playback constraints. Profiling results indicate that modern LSLMs can generate over $100$ tokens per second on GPUs, while naturalistic audio playback typically requires only $12.5$ tokens per second. To exploit this underutilized capacity, the model interleaves speech and reasoning tokens in a fixed proportion, enabling concurrent verbalization and latent inference. Specifically, we constrain the emission rate of spoken tokens to $20$ per second for smooth playback and allocate the remaining generation bandwidth to reasoning. This yields a  $2$ $\mathit{vs.}$ $8$ speech-to-reasoning token ratio, derived directly from the inference budget rather than empirical heuristics. The system is built on the Thinker-Talker architecture~\citep{qwen25omni}, ensuring that interleaved reasoning does not compromise the model’s core language understanding or text-based reasoning performance.

To incentivize the reasoning capabilities of LSLMs under the \emph{“thinking-in-Speaking”} paradigm, we construct a data pipeline and introduce a large-scale dataset, \textsc{Spoken-Math-Problems-3M}, tailored for audio-based mathematical reasoning. Building on prior evidence that mathematical tasks effectively elicit structured cognitive processes in language models, we curate an audio-based dataset of mathematical problems with difficulty comparable to the GSM8K~\citep{gsm8k} benchmark. A key challenge in this setting is \emph{overshooting}, where the verbal output stream advances ahead of the internal reasoning process, leading to premature or hallucinated answers. To address this, we generate two temporally aligned streams for each problem: a fluent, human-readable output sequence and a symbolic, step-by-step reasoning trace. We introduce a prompting strategy that defers substantive content in the output stream while frontloading reasoning steps in the internal stream, thereby establishing a temporal buffer for inference. The resulting streams are tokenized, interleaved, and verified to ensure causal consistency, \emph{i.e.}, no verbal content precedes its logical derivation.
Upon this pipeline, we construct a dataset of 3 million audio-based mathematical reasoning samples by converting a broad collection of publicly available text-based datasets into speech format.

We conduct extensive evaluation on the Spoken-MQA~\citep{SPOKEN-MQA} datasets. Compared with the base model Qwen2.5-Omni-3B, \textsc{Mini-Omni-Reasoner} achieves higher accuracy in both arithmetic (64.9\% → 77.25\% avg, +12.4\%) and reasoning (64.0\% → 68.1\%, +4.1\%), while cutting response length by more than half (42.9 vs. 116.1 words). These results highlight the effectiveness of the proposed \emph{“thinking-in-speaking”} paradigm: unlike Mini-Omni, which sacrifices correctness, or Qwen2.5-Omni-3B, which incurs long delays by verbalizing the entire reasoning chain, our model interleaves reasoning and response tokens but only speaks the latter, thereby preserving correctness while ensuring concise, real-time interaction.

\section{Involving Reasoning in Spoken Dialogue Models}
In this section, we revisit the Thinker-Talker architecture, a state-of-the-art framework for spoken dialogue modeling. We then analyze how to incorporate reasoning into this architecture, illustrating the transition from the conventional \emph{“thinking-before-speaking”} paradigm to our proposed \emph{“thinking-in-speaking”} formulation.

\subsection{Thinker-Talker Pipeline}
The Thinker-Talker framework decouples audio understanding, linguistic inference, and speech synthesis. It consists of three core modules: an audio encoder, a Thinker LLM, and a Talker LLM. Given a raw audio input $\mathbf{x}_{\mathrm{a}}$, the audio encoder first converts it into discrete audio tokens:
$
\mathbf{h}_{1:T}^{\text{a}} = \mathcal{E}_{\text{a}}(\mathbf{x}_{\text{a}}).
$
These tokens, interpreted as linguistic actions, are passed to a Thinker LLM, which autoregressively generates a sequence of response tokens:

\begin{equation}
    \mathbf{t}_{1:N}^{\text{resp}} = \mathcal{T}_{\text{thinker}}(\mathbf{h}_{1:T}^{\text{a}})
\end{equation}

Each generated response token $\mathbf{t}_j^{\mathrm{resp}}$ is immediately mapped into a sequence of audio tokens via the Talker LLM:

\begin{equation}
   \mathbf{z}_j^{\text{a}} = \mathcal{T}_{\text{talker}}(\mathbf{t}_j^{\mathrm{resp}})
\end{equation}

These audio tokens are concatenated to form a continuous stream:

\begin{equation}
    \mathbf{z}_{1:J}^{\text{a}} = \left[ \mathbf{z}_1^{\text{a}} ; \mathbf{z}_2^{\text{a}} ; \dots ; \mathbf{z}_J^{\text{a}} \right]
\end{equation}

To generate audible output, an audio decoder operates on fixed-size sliding windows over this stream. Each audio segment $\hat{\mathbf{x}}_i^{\text{a}}$ is reconstructed from a windowed slice of the audio token stream:

\begin{equation}
    \hat{\mathbf{x}}_i^{\text{a}} = \mathcal{D}_{\text{a}} \left( \mathbf{z}_{s_i : s_i + \ell - 1}^{\text{a}} \right)
\end{equation}

where $s_i$ is the starting index of the $i$-th window and $\ell$ is the predefined audio token segment length. This streaming formulation enables real-time spoken interaction while maintaining modular separation between linguistic reasoning and audio synthesis. It also supports seamless integration of advanced reasoning capabilities within the Thinker module.

\subsection{thinking-before-speaking}
To explore reasoning integration, we start with the \emph{“thinking-before-speaking”} paradigm. Here, the Thinker LLM is augmented to generate a latent reasoning sequence before emitting response tokens. Given the audio token sequence $\mathbf{h}_{1:T}^{\mathrm{a}}$, the Thinker first generates:

$$
\mathbf{t}_{1:M}^{\text{reason}} = \mathcal{T}_{\text{thinker}}(\mathbf{h}_{1:T}^{\text{a}}).
$$

Conditioned on both the audio and reasoning tokens, it then produces the verbal response:

$$
\mathbf{t}_{1:N}^{\text{resp}} = \mathcal{T}_{\text{thinker}}(\mathbf{h}_{1:T}^{\text{a}}, \mathbf{t}_{1:M}^{\text{reason}}).
$$

In this case, we consider two decoding strategies for the Talker LLM depending on how it handles reasoning tokens $\mathbf{t}_{1:M}^{\mathrm{reason}}$ and response tokens $\mathbf{t}_{1:N}^{\mathrm{resp}}$.

\noindent\textbf{Full Verbalization.} In this approach, both reasoning and response tokens are converted into audio:
\begin{equation}
   \hat{\mathbf{x}}_{1:(M+N)}^{\text{a}} = \mathcal{D}_{\text{a}}(\mathcal{T}_{\text{talker}}([\mathbf{t}_{1:M}^{\text{reason}}; \mathbf{t}_{1:N}^{\text{resp}}])). 
\end{equation}

This produces a complete narration including reasoning and answer, but requires the listener to hear through reasoning content before the actual answer, introducing potential cognitive overload.

\noindent\textbf{Silent Reasoning.} Alternatively, the Talker LLM remains silent during reasoning token generation and only begins decoding when the first response token segment is available:

\begin{equation}
    \hat{\mathbf{x}}_t =
\begin{cases}
\text{silent}, & \text{if } \mathbf{t}_i \in \mathbf{t}_{1:M}^{\text{reason}} \\
\mathcal{D}_{\text{audio}}\left( \mathcal{T}_{\text{talker}}(\mathbf{t}_i) \right), & \text{if } \mathbf{t}_i \in \mathbf{t}_{1:N}^{\text{resp}},
\end{cases}
\end{equation}
where $t_i$ denotes a token generated by the Thinker LLM. This strategy ensures that only essential information is verbalized, improving clarity and efficiency, though it incurs a first-token delay due to the reasoning phase.

\subsection{thinking-in-speaking}
To address the trade-off between reasoning depth and response latency, we introduce a novel \emph{“thinking-in-speaking”} paradigm that interleaves reasoning and response generation. Unlike the conventional thinking-before-speaking approach, which delays response until reasoning is complete, our method enables the Thinker LLM to alternate between generating $p$ response tokens and $q$ reasoning tokens:

\begin{equation}
\mathbf{t}_{1:(p+q)\cdot K} = \bigcup_{i=1}^{K} \left\{ \mathbf{t}_{(i-1)(p+q)+1}^{\text{resp}}, \dots, \mathbf{t}_{(i-1)(p+q)+p}^{\text{resp}}, \;\mathbf{t}_{(i-1)(p+q)+p+1}^{\text{reason}}, \dots, \mathbf{t}_{i(p+q)}^{\text{reason}} \right\}.
\end{equation}

During token prediction, the Talker LLM operates in a selective manner: it converts only the response segments into audio while remaining silent for the reasoning tokens. Compared to \emph{thinking-before-speaking}, which waits for all $\mathbf{t}^{\text{reason}}$ to finish before any speech is produced, our interleaved generation scheme allows real-time response streaming while reasoning is still in progress.

Once a response token $\mathbf{t}_{i}^{\text{resp}}$ is generated, it is passed through the Talker for real-time conversion into speech:

\begin{equation}
    \hat{\mathbf{x}}_{i}^{\text{a}} = \mathcal{D}_{\text{audio}} \left( \mathcal{T}_{\text{talker}}(\mathbf{t}_{i}^{\text{resp}}) \right).
\end{equation}

This strategy exploits the empirical observation that token generation in autoregressive LLMs is significantly faster than real-time audio rendering. Thus, response segments can be emitted promptly while reasoning continues in the background, enabling continuous, low-latency interaction.

The $(p, q)$ ratio serves as a tunable parameter, balancing reasoning granularity with responsiveness, and can be adapted based on model throughput and deployment constraints. The full design and implementation of this \emph{“thinking-in-speaking”} pipeline are elaborated in the following section.

\section{Mini-Omni-Reasoner}
Grounded in our novel \emph{“thinking-in-speaking”} paradigm, we introduce Mini-Omni-Reasoner, a reasoning-involved framework for real-time spoken dialogue. This framework builds upon the Thinker-Talker architecture by integrating token-level interleaving of reasoning and response generation, enabling both zero-latency audio output and improved inference quality. \textsc{Mini-Omni-Reasoner} is particularly suited for complex mathematical and logical tasks, where structured reasoning is critical to response accuracy. To support this capability, we develop (i) an interleaved generation pipeline that temporally aligns internal reasoning and verbal output, (ii) a large-scale spoken math dataset derived from symbolic problem sets and rendered via TTS, and (iii) a progressive training strategy that transitions the model from standard dialogue modeling to reasoning-aware audio generation.

\begin{figure}[t]
    \centering
    \includegraphics[width=1\linewidth]{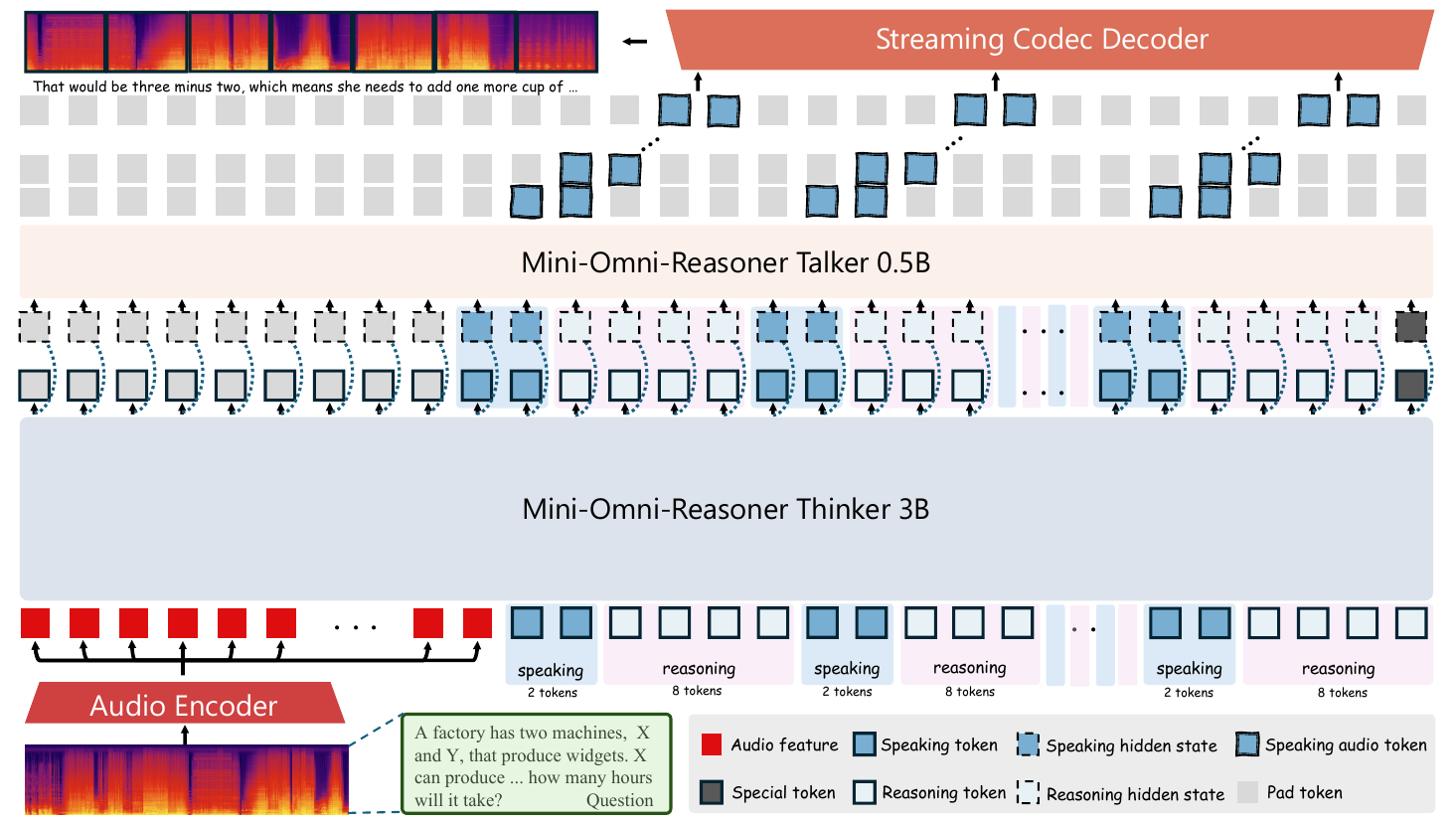}
    \caption{\textbf{Overview of the Mini-Omni-Reasoner.} Given a raw audio instruction, the audio encoder transforms it into a sequence of audio tokens embedded in language space. These tokens are used to pre-fill the Thinker LLM, initializing its context for autoregressive generation. The Thinker then generates an interleaved sequence of answer and reasoning tokens. Unlike conventional approaches that emit only answer tokens, our interleaved formulation enables the model to engage in internal reasoning while maintaining uninterrupted response generation. The answer tokens are streamed into the Talker module and decoded into speech in real time via a codec decoder, whereas the reasoning tokens remain silent and guide the generation process. This \emph{“thinking-in-speaking”} formulation enables real-time, zero-latency audio interaction while preserving the model’s reasoning capabilities.
    }
    \label{fig:2-framework}
\end{figure}

\subsection{The Pipeline of \textsc{Mini-Omni-Reasoner}}
As illustrated in Figure~\ref{fig:2-framework}, \textsc{Mini-Omni-Reasoner} adopts a hierarchical Thinker–Talker architecture that supports real-time spoken reasoning. The core innovation lies in the Thinker, which integrates token-level internal reasoning with externally observable response generation, embodying the \emph{“thinking-in-speaking”} paradigm. The Thinker comprises three main components: an audio encoder, an audio adapter, and a language model. Given a raw audio input, the audio encoder, pretrained on large-scale audio datasets, extracts high-quality audio features. These features are then projected into a linguistic space through the audio adapter and incorporated as prefix embeddings into the language model. We initialize the Thinker with \texttt{Qwen2.5-Omni-3B}. This module is then trained using the interleaved token-level objective described in Section~\ref{subsec-training_setup} and is frozen after training to preserve its reasoning ability. The Talker module is a compact model with the same architecture as the Thinker but trained separately from scratch. It learns to predict audio tokens using the SNAC tokenizer~\citep{snac}, conditioned on the Thinker’s response tokens. This separation of responsibilities enables the Talker to generate fluent speech outputs while leveraging the Thinker’s reasoning capacity. This hierarchical design allows for a clean modularization of cognitive functions, supporting accurate reasoning and low-latency, naturalistic spoken interaction in real time.

\noindent\textbf{Token-Level Thinking-in-Speaking.}
We introduce how \textsc{Mini-Omni-Reasoner} implements token-level \emph{“thinking-in-speaking”} within the Thinker module. Conventional large language models, such as OpenAI-o4~\citep{openai-o4} and DeepSeek-R1~\citep{deepseek-r1}, typically adopt a \emph{“thinking-before-speaking”} strategy, generating a full reasoning trace before emitting any response tokens. Despite not following this sequential reasoning-first approach, our base model \texttt{Qwen2.5-Omni}~\citep{qwen25omni} still consolidates the complete reasoning chain into its speech output, resulting in long response latency, verbose reasoning, and a suboptimal user experience.
 To address this issue, \textsc{Mini-Omni-Reasoner} employs an \emph{interleaved generation strategy}. The model alternates between producing outward-facing response tokens and inward-facing reasoning tokens. We adopt a fixed interleaving ratio of $2$ $\mathit{vs.}$ $8$, meaning the model emits two response tokens followed by eight reasoning tokens per cycle. This design ensures continuous and natural speech while preserving sufficient internal reasoning for reliable decision-making.

This mechanism relies on two key components. First, the interleaving ratio controls the trade-off between conversational fluency and reasoning depth. Second, special control tokens are introduced to explicitly demarcate reasoning and response segments during both training and inference. Together, these design choices support fluid, reasoning-aware spoken interactions.

\underline{\emph{Design of Reasoning–Response Token Ratio.}}
The interleaving ratio between reasoning and response tokens is critical for balancing latency, reasoning quality, and controllability. We adopt a $2$ $\mathit{vs.}$ $8$ setting for three main reasons:
(1) \textbf{Controllability}: short response blocks help avoid premature verbal output before adequate reasoning has been performed.
(2) \textbf{Reasoning Capacity}: the $2$ $\mathit{vs.}$ $8$ ratio ensures that the model dedicates four times more tokens to internal reasoning than to speech, enabling deeper deliberation.
(3) \textbf{Real-Time Compatibility}: a 3B-scale model typically generates around 100 tokens per second on a standard GPU. Under this setting, it produces approximately 20 response tokens per second (roughly five words), which is sufficient for smooth speech synthesis. Empirically, this configuration offers a strong trade-off between responsiveness and reasoning robustness, and serves as the default setting for \textsc{Mini-Omni-Reasoner}.

\underline{\emph{Control Token Design.}}
Special tokens play an essential role in maintaining the alternation between reasoning and response streams. We evaluate three strategies:
(1) \textbf{No explicit marker}: the model is trained without explicit boundaries, relying on pattern learning. This proves unstable as the model drifts from the intended pattern.
(2) \textbf{Explicit markers with loss weighting}: split tokens are inserted and emphasized during training. This leads to unstable placement and poor alignment.
(3) \textbf{Masked markers}: split tokens are inserted but masked from the loss computation during training. This approach avoids overfitting and proves most effective. During inference, we manually insert these markers to guide generation. \textsc{Mini-Omni-Reasoner} adopts the masked token strategy and further appends eight padding tokens after each reasoning block. This padding stabilizes alignment for the downstream Talker and ensures consistent adherence to the $2$ $\mathit{vs.}$ $8$ interleaving ratio.

\subsection{Spoken-Math-Problem Dataset}

\begin{figure}[t]
    \centering
    \includegraphics[width=1\linewidth]{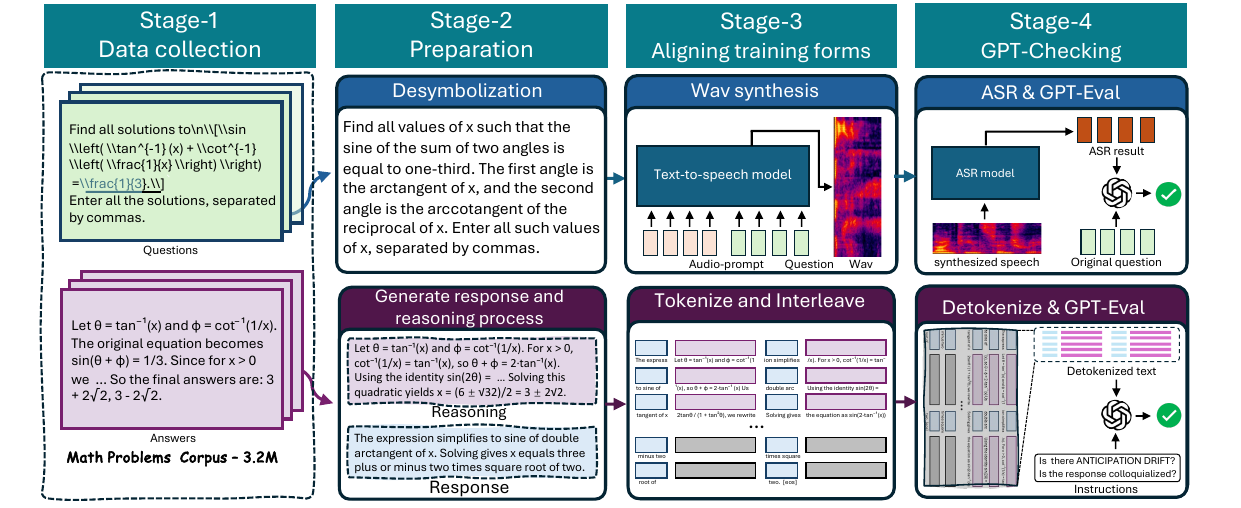}
    \caption{\textbf{Pipeline for Constructing the Spoken-Math-Problem-3M Dataset.} We first aggregate a large-scale dataset of math problems from publicly available text-based datasets. These problems are reformulated into spoken-style natural language to align with audio-based interaction settings. The reformulated prompts are synthesized into speech via a TTS tool. Finally, a GPT-based verification stage is applied to ensure fluency, coherence, and semantic fidelity of the generated audio-text pairs.
    }
    \label{fig:data-construction}
\end{figure}

A key prerequisite for enabling reasoning-in-speaking within the \textsc{Mini-Omni-Reasoner} framework is the construction of high-quality aligned training data. In this section, we first identify a central challenge: anticipation drift, a phenomenon in which the speaking process outpaces the underlying reasoning trajectory, resulting in logical misalignment and degraded inference quality. To mitigate this issue, we design a structured data synthesis pipeline that tightly couples reasoning sequences with temporally coherent spoken outputs. Finally, we present a quantitative analysis of the resulting training dataset, detailing its scale, composition, and reasoning complexity.

\textbf{thinking-in-speaking Data Formulation.} The interleaved \emph{“thinking-in-speaking”} paradigm enables real-time verbal interaction by alternating between reasoning and response tokens at a fine-grained level. However, this generation scheme introduces a semantic alignment challenge: ensuring that each response token is grounded in sufficient prior reasoning. While the total number of reasoning tokens typically exceeds that of response tokens, the interleaved token order does not guarantee that reasoning precedes its corresponding response content. This misalignment is particularly prominent at the early stages of generation, where response tokens may appear before any substantive reasoning has been produced. Such cases violate the intended \emph{“thinking-before-speaking”} principle in semantic terms, where verbal content should logically follow from internal deliberation. To address this issue during data construction, we propose a two-stage strategy that formulates the \emph{“thinking-in-speaking”} training sequences to ensure semantic precedence of reasoning. 

First, we introduce an asynchronous alignment scheme inspired by natural human dialogue, in which speakers often begin with light contextualization—such as greetings or conversational scaffolding—before delivering reasoning-grounded content. To emulate this structure, we impose differentiated constraints on the two streams: the reasoning sequence must begin immediately with substantive logical inference, avoiding redundant or introductory tokens; the response sequence is encouraged to adopt a delayed-onset structure, starting with softening phrases or contextual cues before expressing content derived from the reasoning trace. This offset establishes semantic precedence of reasoning while accommodating interleaved generation during inference.

Second, we introduce a sequence-level verification process as a post-processing step. Each reasoning–response pair is tokenized and reassembled using the $2$ $\mathit{vs.}$ $8$ interleaving ratio to simulate the model's generation order. The resulting hybrid sequence is detokenized to approximate the actual spoken output. A GPT-based evaluator then checks for (1) premature appearance of reasoning content and (2) semantic and temporal consistency between the reasoning and response streams. Only examples that pass this screening are retained for training. This procedure ensures that the final dataset faithfully adheres to the intended \emph{“thinking-before-speaking”} logic, even under token-level interleaving, and supports accurate and coherent reasoning in real-time spoken interaction.

\textbf{Dataset Construction Pipeline.}  To support the training of reasoning-capable spoken language models, we construct a large-scale pretraining dataset through a structured four-stage pipeline comprising data collection, data preparation, training format alignment, and semantic verification. An overview of this process is illustrated in Figure~\ref{fig:data-construction}.

We initiate the pipeline by curating a diverse and scalable dataset centered on math word problems, a domain that demands both abstract reasoning and verbal clarity. Rather than relying on limited speech-based data, we resample from high-quality text-based math QA datasets, resulting in a corpus of 3M instances, denoted as \textsc{Spoken-Math-Problems-3M}. This collection offers broad coverage of mathematical reasoning styles and ensures sufficient volume to support the training of large-scale models.

In the data preparation phase, we process questions and answers independently. Each question is passed through a rewriting module to generate two forms: one that retains the original syntax and another that reformulates the prompt into a more conversational and speech-friendly style. This dual-format design enables flexible downstream speech synthesis. For the answers, we adopt the reasoning-before-response formulation described in above. Each instance is decomposed into a symbolic reasoning trace followed by a concise spoken response. The reasoning portion is constructed to reflect logical deduction steps, while the response aims for accessibility and listener fluency. The relative length of the reasoning is maintained at roughly twice that of the response to preserve sufficient cognitive grounding.

To convert the processed data into training form, we synthesize the rewritten questions into waveform audio via the CosyVoice2-0.5B TTS model~\citep{cosyvoice2}, which provides high-fidelity audio suitable for instruction-like prompts. The paired reasoning and response texts are then tokenized and interleaved at a fixed  $2$ $\mathit{vs.}$ $8$ ratio, mimicking the token-level alternation pattern required for \emph{“thinking-in-speaking”}. This interleaving ensures that reasoning content is sufficiently introduced before each response segment, enabling better semantic alignment between internal computation and verbal output.

The final stage, GPT-based verification, ensures that the generated output does not suffer from overshooting, a failure mode in which answer content appears before the corresponding reasoning process is complete. This phenomenon compromises logical coherence and often leads to hallucinated or ungrounded responses. To detect and eliminate such cases, the interleaved token sequences (constructed using the  $2$ $\mathit{vs.}$ $8$ ratio) are detokenized into natural language and passed through a semantic verification model. This model checks that every response token is appropriately supported by preceding reasoning, thereby preserving the alignment between internal deliberation and verbal output.

\subsection{Training Methodology}
\begin{figure}[t]
    \centering
    \includegraphics[width=1\linewidth]{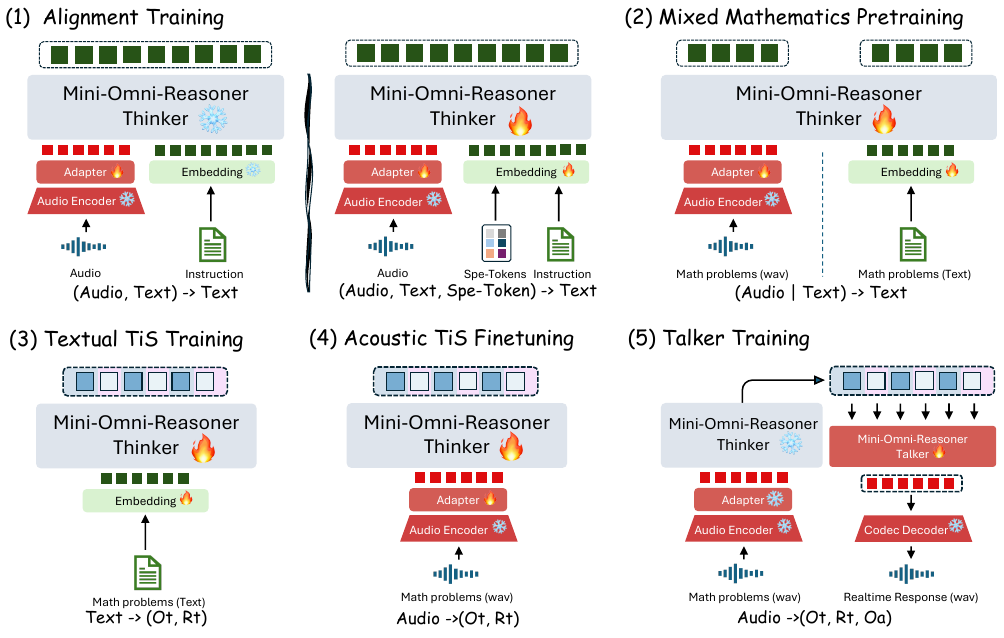}
    \caption{\textbf{Training Pipeline of \textsc{Mini-Omni-Reasoner}.} We initialize the \textsc{Mini-Omni-Reasoner} with a Thinker-Talker architecture, where the Thinker is a pretrained large language model and the Talker is randomly initialized.  To enable interleaved reasoning and speaking, we progressively adapt the system through a multi-stage training process. The model learns to alternate between generating answer tokens for real-time speech synthesis and reasoning tokens for internal inference, forming a tightly coupled loop between reasoning quality and audio responsiveness.
    }
    \label{fig:4-training_stages}
\end{figure}
Training the proposed Mini-Omni-Reasoner requires a carefully staged pipeline, as it introduces both a customized model architecture and a novel output formulation. To ensure stable convergence and effective transfer of reasoning capabilities from text to speech, we decompose the training process into five progressively more complex stages. Each stage is designed to first preserve or enhance reasoning within the textual modality, then align it with the speech modality. An overview of this process is presented in Figure~\ref{fig:4-training_stages}.
\textbf{Stage 1: Alignment Training.} 
We initialize Mini-Omni-Reasoner from Qwen2.5-Omni-3B and resolve architectural inconsistencies to ensure compatibility. This includes adapting to implementation differences such as RoPE variants. In this stage, we first fine-tune only the audio adapter using speech QA and dialogue datasets, while freezing the rest of the model. This bridges the interface between the speech encoder and the LLM backbone. Subsequently, we unfreeze all components except the audio encoder to adapt to newly introduced special tokens, which are embedded into the tokenizer’s reserved ID space. This enables the model to function seamlessly under the customized token format.
\textbf{Stage 2: Mixed Mathematical Pretraining.} 
With the model aligned, we enhance its mathematical reasoning ability as a prerequisite for interleaved generation. To isolate representation learning from paradigm learning, we pretrain the model on standard \emph{“thinking-before-speaking”} datasets that include both speech and text forms. This ensures strong reasoning capability and data alignment before introducing the complexity of token-level interleaving.
\textbf{Stage 3: Textual Thinking-in-Speaking.} 
We begin paradigm-specific training using the text modality, which is easier to model than speech. The model learns to alternate between reasoning tokens and response tokens within a single sequence. During this stage, we update only the language model parameters to focus solely on internalizing the interleaved reasoning-response structure, without introducing acoustic variability.
\textbf{Stage 4: Acoustic Thinking-in-Speaking.} 
Having established interleaved generation in the text domain, we transition to spoken inputs. Textual queries are replaced with audio, and the audio encoder is fine-tuned while the LLM remains fixed. This allows the model to maintain reasoning-augmented generation when conditioned on speech, effectively transferring the reasoning paradigm across modalities.
\textbf{Stage 5: Talker Training.}
In the final stage, we enable fluent speech generation by training the talker module. The entire “thinker” component—comprising all modules trained in the prior stages—is frozen. We train only the talker to synthesize speech from the interleaved outputs, ensuring that spoken responses remain natural and coherent while preserving the logical grounding developed earlier.

\section{Experiments}

\subsection{Training Setup \label{subsec-training_setup}}

Our training process build upon the mini-omni codebase, where we reconstruct the foundational model architecture from scratch. Specifically, we adopt the Qwen2.5-Omni encoder module as the audio encoder to extract speech features, and introduce a single linear adapter layer to bridge the audio encoder and the language model. The core language model is based on Qwen2.5-3B, which together with the encoder and adapter forms the \textsc{Mini-Omni-Reasoner} framework. To ensure parameter alignment and stable convergence, all model components are initialized from the corresponding modules of the pre-trained Qwen2.5-Omni-3B checkpoint. Training is conducted on 32 NVIDIA H100 GPUs, leveraging data parallelism for efficiency. We train on a large-scale dataset containing 3 million samples, running for 6 full epochs with a batch size of 64. The learning rate follows a cosine decay schedule, with the maximum learning rate set to 2e-4.

\subsection{Benchmark}
We use the Spoken-MQA~\citep{SPOKEN-MQA} benchmark to comprehensively evaluate spoken mathematical reasoning ability. Spoken-MQA consists of two main categories: Arithmetic and Contextual Reasoning. The \textbf{Arithmetic} category tests fundamental numerical operations including addition, subtraction, multiplication, and division with both integer and decimal numbers, focusing on direct computation with minimal contextual knowledge. The \textbf{Contextual Reasoning} category contains everyday word problems requiring interpretation of short narratives and arithmetic reasoning, further divided into single-step problems from AddSub~\citep{add1,add2} and SingleOp~\citep{singleop} datasets and multi-step problems from GSM8K~\citep{gsm8k} and SVAMP~\citep{patel2021nlp}, reflecting increasing complexity and sensitivity to linguistic variations. The number of samples for each category is shown in Table~\ref{tab:task_statistics}.
\begin{table}[h]
\centering
\caption{Spoken-MQA sub-task statistics.}
\label{tab:task_statistics}
\begin{tabular}{lcc}
\toprule
\textbf{Sub-task} & \textbf{Category} & \textbf{\#Samples} \\
\midrule
short\_digit       & Arithmetic            & 118  \\
long\_digit        & Arithmetic            & 155  \\
single\_step\_reasoning & Contextual Reasoning   & 594  \\
multi\_step\_reasoning  & Contextual Reasoning   & 1402 \\
\bottomrule
\end{tabular}
\end{table}

\subsection{Baselines} 
We compare our proposed model against three categories of baselines to comprehensively evaluate its effectiveness and performance ceiling.
\begin{itemize}[leftmargin=*]
    \item  \textbf{Cascade Models}. We include cascade models to validate the benchmark’s reliability and establish an upper bound based on aligned text-to-text models. Specifically, the cascade approach consists of Whisper-v3-large~\citep{whisper} for speech recognition, followed by text processing using two advanced language models: Qwen2.5-Instruct-7B~\citep{qwen25} and Qwen2.5-Math-7B-Instruct~\citep{qwen25math}.

    \item  \textbf{Speech Models.} We benchmark against a variety of mainstream dialogue models including SLAM-Omni~\citep{slam-omni}, Mini-Omni~\citep{xie2024mini}, Moshi~\citep{moshi}, LLaMA-Omni~\citep{llama-omni}, Freeze-Omni~\citep{freezeomni}, Qwen2-Audio-Instruct~\citep{qwen2-audio}, and Qwen2.5-Omni-7B~\citep{qwen25omni}. For Qwen2-Audio-Instruct and Qwen2.5-Omni-7B, we further leverage prompt engineering to enable step-by-step reasoning (“think step by step”) during inference, allowing a deeper comparison of reasoning capabilities.

    \item  \textbf{Foundation Model.} We finally compare with the base model Qwen2.5-Omni-3B itself, evaluating its performance under both the standard setting and the “think step by step” mode. This comprehensive baseline setup enables us to analyze the contribution of our model against both pipeline-based and end-to-end reasoning-capable models across multiple inference strategies.

\end{itemize}

\begin{table}[t]
\centering
\caption{Spoken-MQA results (\%). Best per column in bold. Models with * indicate that the prompt includes “please think step by step.”}
\resizebox{\linewidth}{!}{
\begin{tabular}{llccccccc}
\toprule
\multirow{2}{*}{Models} & \multirow{2}{*}{Size} & \multicolumn{3}{c}{Arithmetic} & \multicolumn{3}{c}{Reasoning} & \multirow{2}{*}{Avg} \\
\cmidrule(lr){3-5} \cmidrule(lr){6-8}
& & Short & Long & Avg & Single & Multi & Avg & \\
\midrule
\multicolumn{9}{l}{\textit{Cascade}} \\
Whisper-Qwen2.5-7B-Instruct       & 7B & - & - & 70.0    & - & - & \underline{72.5} & \underline{72.2}\\
Whisper-Qwen2.5-Math-7B-Instruct  & 7B & - & - &  \underline{77.3}   &-  & - &  \underline{86.7} & \underline{85.6} \\
\midrule
\multicolumn{9}{l}{\textit{Conversational Models}} \\
SLAM-Omni        & 0.5B & 0.0 & 0.0 &  & 0.8 & 1.4 & 1.22 & 1.1 \\
Moshi            & 7B & 0.0 & 0.0 &  & 0.2 & 0.2 & 0.2 & 0.2 \\
LLaMA-Omni       & 7B & 40.0 & 11.0 & 23.5 & 29.5 & 10.5 & 16.2 & 16.8 \\
Mini-Omni         & 7B & 5.0 & 2.3 & 3.5  & 0.8 & 1.9 & 1.6 & 1.7 \\
Freeze-omni        & 7B & 43.0 & 14.5 & 26.8 & 69.0 & 19.8 & 34.4 & 33.3 \\
GLM-4-Voice       & 9B & 40.0 & 22.5 & 30.1  & 54.4 & 28.5  & 36.2 & 35.3 \\
Qwen2-Audio-7B-Instruct & 7B & 61.0 & 39.3 & 48.7 & 56.3 & 21.2 & 31.7 & 33.7 \\
Qwen2-Audio-7B-Instruct* & 7B & 43.0 & 31.2 & 36.3 & 55.4 & 22.5 &  32.3 & 32.7 \\
Qwen2.5-Omni-7B & 7B & 90.0 & 49.1 & 66.8 & 84.9 & \underline{71.0} & \underline{75.1} & \underline{73.8} \\
Qwen2.5-Omni-7B* & 7B & 83.0 & 45.1 & 61.5 & 85.2 & \underline{71.5} & \underline{75.6} & \underline{73.6} \\
\midrule
\multicolumn{9}{l}{\textit{Baseline}} \\
Qwen2.5-Omni-3B & 3B & 87.0 & 48.0 & 64.9 & 81.8 & 56.4 & 64.0 & 63.7 \\
Qwen2.5-Omni-3B* & 3B & 84.0 & 43.3 & 60.1 & 81.5 & 57.1 & 64.4 & 63.6 \\
\midrule
\multicolumn{9}{l}{\textit{Ours}} \\
Mini-Omni-Reasoner & 3B & \textbf{92.9} & \textbf{66.1} & \textbf{77.25} & \textbf{85.9} & \textbf{60.5} & \textbf{68.1} & \textbf{68.6} \\
\bottomrule
\end{tabular}
}
\label{tab:spoken_mqa}
\end{table} 

\subsection{Performance on Spoken-MQA}
Table~\ref{tab:spoken_mqa} presents the Spoken-MQA results, separating arithmetic computation from contextual reasoning. The analyses are as follows:

\textit{(1) Arithmetic performance.} \textsc{Mini-Omni-Reasoner} achieves the highest short-form score (92.9\%), outperforming all cascade and conversational models, including the cascade-based Whisper-Qwen2.5-Math-7B-Instruct (77.3\%) and the best open-source conversational baseline, Qwen2.5-Omni-3B (87.0\%). The model also leads on long-form arithmetic (66.1\%), where most conversational systems struggle (e.g., Mini-Omni at 2.3\%, LLaMA-Omni at 11.0\%), indicating superior robustness in extended spoken numerical computation. Overall, it attains a category average of 77.25\%, exceeding the strongest cascade model by +4.0\% and surpassing the best conversational alternative by +3.2\%, reflecting substantial gains in both precision and generalization. 

\textit{(2) Reasoning performance.} In the \textbf{Reasoning} category, \textsc{Mini-Omni-Reasoner} again establishes new state-of-the-art performance among open-source systems. It achieves 85.9\% on single-step reasoning tasks, marginally higher than Qwen2.5-Omni-7B (84.9\%) and cascade-based Whisper-Qwen2.5-Math-7B-Instruct (86.7\%), while maintaining a competitive 60.5\% on multi-step reasoning—substantially above the majority of conversational models (e.g., GLM-4-Voice at 28.5\%, Qwen2-Audio-7B-Instruct at 21.2\%). The resulting average of 68.1\% in this category surpasses the best baseline (64.0\%) by +4.1\%, indicating strong interpretive ability across both simple and compositional reasoning. 

\textit{(3) Overall.} The results demonstrate that the proposed training paradigm of \textsc{Mini-Omni-Reasoner} is effective and does not compromise the model’s generalization capability. It matches or exceeds the performance of 7B-scale and cascade models in arithmetic, showing high reliability. On challenging tasks such as multi-step reasoning and long-digit computation, it consistently outperforms the 3B baseline (Qwen2.5-Omni-3B) by a clear margin. While its performance on these tasks is slightly lower than some 7B models, we attribute this gap primarily to model size rather than limitations in the training strategy.

\subsection{Response Latency Comparison}
\begin{figure}[!t]
    \centering
    \includegraphics[width=1\linewidth]{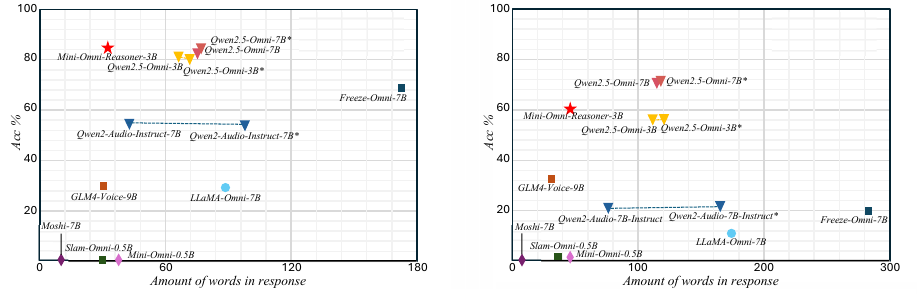}
    \caption{\textbf{Performance and Spoken Response Length Comparison on the Spoken-MQA.} \emph{Left:} a single-step resoning math problem (simple); \emph{right:} a multi-step reasoning problem (hard). \textsc{Mini-Omni-Reasoner}-3B outperforms existing 3B-scale models and achieves comparable performance to 7B models, while generating shorter spoken responses. This demonstrates its ability to deliver concise and informative answers, reducing audio latency without compromising reasoning quality.
    }
    \label{fig:figure5-lengths}
\end{figure}

Figure~\ref{fig:figure5-lengths}  evaluates model performance and response characteristics on Spoken-MQA, encompassing single/multi-word understanding and reasoning tasks. \textsc{Mini-Omni-Reasoner}, fine-tuned from Qwen2.5-Omni-3B, exhibits exceptional efficiency: it achieves 85.9\% in single-task reasoning—surpassing Qwen2.5-Omni-3B (81.8\%) and even matching Qwen2.5-Omni-7B* (85.2\%)—while delivering 60.5\% in multi-task reasoning, outperforming GLM4-Voice (28.5\%) and Qwen2-Audio variants (21.2\%-22.5\%) by significant margins. Notably, despite trailing Qwen2.5-Omni-7B (71.0\%) on complex multi-task reasoning, \textsc{Mini-Omni-Reasoner}'s user-perceived response length (42.9 words) is less than 50\% of Qwen2.5-Omni-7B's (116.13 words).

This efficiency stems from its \emph{“thinking-in-speaking”} paradigm: while total generation length (think + response) doubles that of Qwen2.5-Omni-3B models, interleaved generation reduces user-audible content to 25\% of total length, enabling faster information delivery. In contrast, larger models like Qwen2.5-Omni-7B* (118.27 words) and Freeze-Omni (283.86 words) produce excessively verbose responses without proportional accuracy gains. Open-source alternatives such as SLAM-Omni (1.4\% multi-task accuracy) and Moshi (0.2\%) demonstrate minimal reasoning capabilities, while LLaMA-Omni (10.5\%) and Mini-Omni (1.9\%) lag significantly despite varying output lengths. These results highlight \textsc{Mini-Omni-Reasoner}'s unique balance of reasoning prowess and communication efficiency.

\subsection{Training Analysis}

\begin{figure}[h]
    \centering
    \includegraphics[width=1\linewidth]{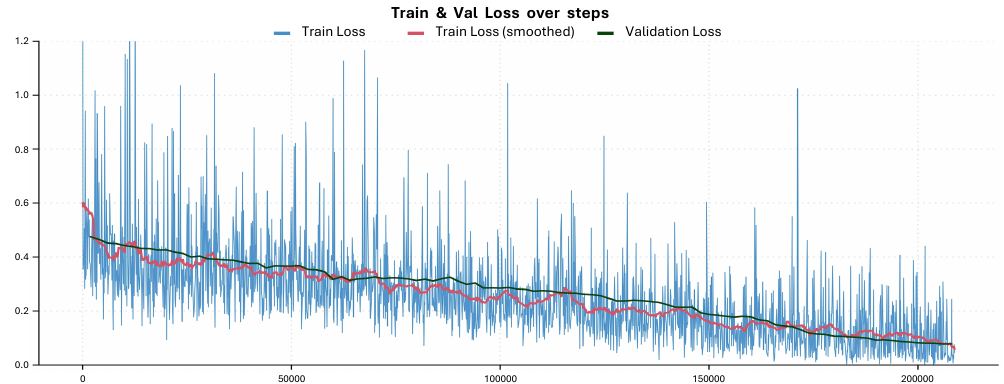}
    \caption{Training and Validation Loss Curves.}
    \label{fig:6-loss}
\end{figure}

We initially harbored concerns that the constant switching between the “speak” and “think” modes would pose immense training challenges, potentially causing chaos in data distribution and hindering model convergence. However, the loss curve depicted in the figure tells a reassuring story, as showned in Figure~\ref{fig:6-loss}. The training loss initiates at approximately 0.6, and as training progresses through over 200,000 steps, it exhibits a distinct and consistent downward trajectory, ultimately stabilizing at around 0.1. The smoothed training loss, which mitigates the inherent fluctuations of the raw training loss, also shows a steady descent, starting from roughly 0.5 and tapering off to about 0.1 as well. Meanwhile, the validation loss follows a highly similar pattern: beginning at around 0.5, it gradually decreases in tandem with the training loss and, in the later stages, aligns closely with the smoothed training loss. Such a smooth and well-behaved loss curve, with both training and validation losses showing coherent and stable reduction without signs of divergence or erratic behavior, strongly demonstrates that the training approach of \textsc{Mini-Omni-Reasoner} is reasonable and effective. It successfully overcomes the potential hurdles introduced by the alternating “speak” and “think” mechanism, validating the viability of our training strategy.

\subsection{Case Studies}
\begin{figure}[!h]
    \centering
    \includegraphics[width=1\linewidth]{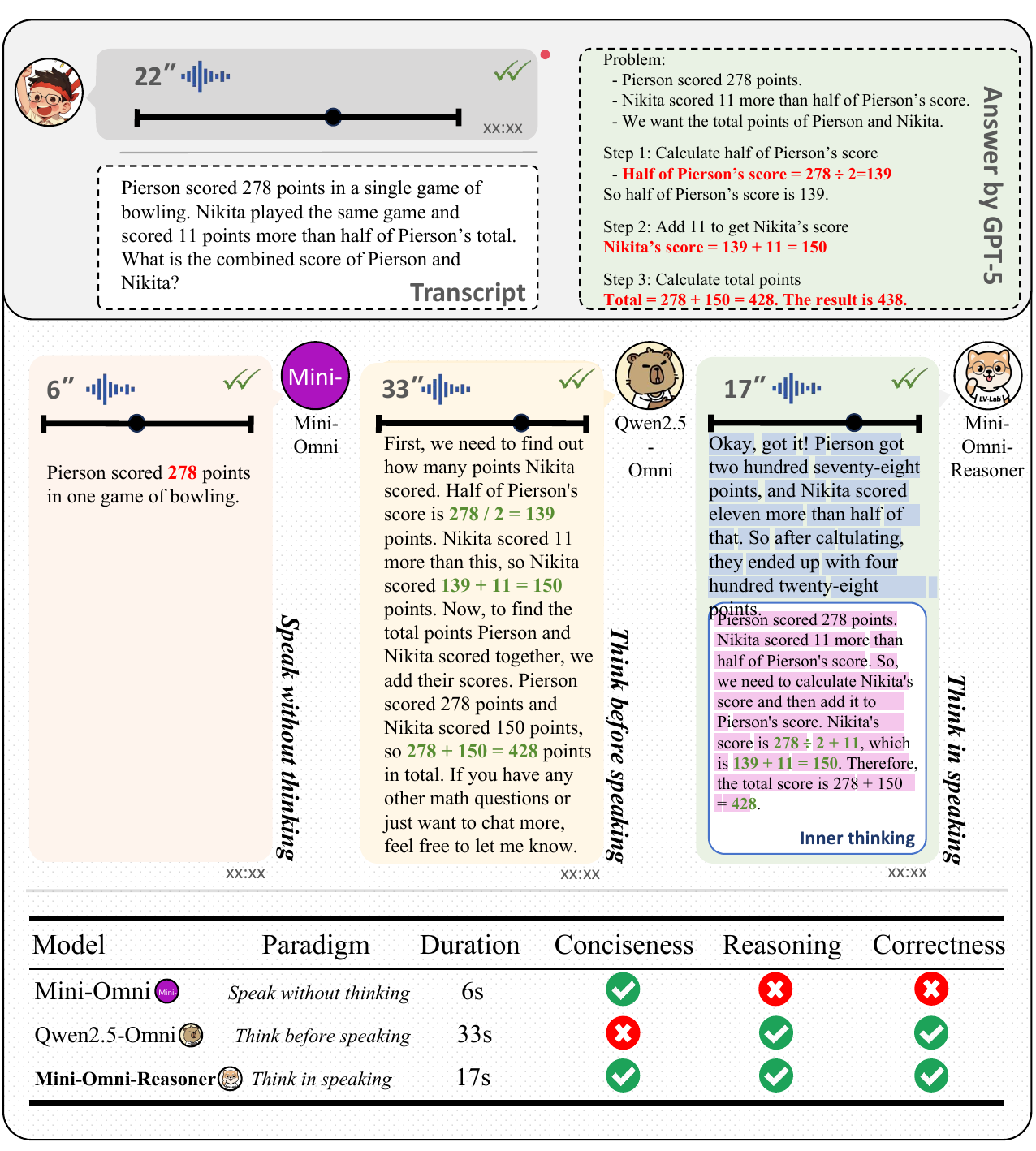}
    \caption{\textbf{Comparison of three speech model paradigms.} Early models like Mini-Omni perform simple dialogue with \textit{speaking-without-thinking}. Qwen2.5-Omni, built on Thinker-Talker, supports reasoning but verbalizes the full chain, causing long and delayed outputs. Mini-Omni-Reasoner adopts \textit{thinking-in-speaking}, delivering high-quality reasoning while keeping responses concise.}
    \label{fig:7-cases}
\end{figure}

\newpage
In this section, we provide a case study to compare the effectiveness of the proposed \emph{“thinking-in-speaking”} paradigm against three alternative end-to-end speech models, as illustrated in Figure~\ref{fig:7-cases}. Specifically, we consider: (i) Mini-Omni, which represents \emph{“speaking-without-reasoning”} by directly mapping inputs to spoken answers without any reasoning traces, (ii) Qwen2.5-Omni-3B, which follows a \emph{“thinking-before-speaking”} strategy by conducting full reasoning in the speech domain such that the entire reasoning trajectory is synthesized into speech, and (iii) \textsc{Mini-Omni-Reasoner}, our model, which adopts the \emph{“thinking-in-speaking”} paradigm by interleaving reasoning tokens and response tokens, while only synthesizing the response into speech. The results reveal clear differences across paradigms. Models like Mini-Omni, despite achieving highly efficient responses, consistently fail to ensure correctness due to the absence of reasoning. In contrast, Qwen2.5-Omni-3B successfully produces accurate answers by synthesizing its complete reasoning process, but this leads to extremely long spoken outputs, requiring tens of seconds for users to obtain the final answer. \textsc{Mini-Omni-Reasoner} achieves a favorable balance: although it generates more reasoning tokens than Mini-Omni, it drastically reduces response latency by using concise phrases (e.g., “after calculating”) to summarize the computation, thereby halving the overall response time while preserving correctness. Finally, we summarize the comparison in the table at the bottom, which demonstrates that \emph{“thinking-in-speaking”} combines the correctness of reasoning-based paradigms with the efficiency of direct-answering approaches.

\section{Related work}
\textbf{Speech LLMs.}  Traditional speech dialogue systems rely on ASR, text generation, and TTS pipelines, which introduce substantial latency. Large audio language models, such as \textsc{SpeechGPT}~\citep{speechgpt} and \textsc{Qwen2-Audio}~\citep{qwen2audio}, partially address this by directly processing speech and generating outputs, and GPT-4o further enables ultra-low-latency end-to-end interaction. \textsc{Mini-Omni}~\citep{xie2024mini} extends this line by introducing a text-guided paradigm to generate speech tokens in parallel with text, bridging the gap between text generation and TTS, and represents the first open-source end-to-end solution. Similar works include \textsc{Freeze-Omni}~\citep{freezeomni}, \textsc{LLaMA-Omni}~\citep{llama-omni}, and dual-stream full-duplex models like \textsc{Moshi}. While early speech LLMs focused on simple dialogue, models such as \textsc{GLM4-Voice}~\citep{glm-voice} and \textsc{Qwen2.5-Omni}~\citep{qwen25omni} achieve near-text-level alignment, yet still follow the text-guided paradigm introduced by Mini-Omni, producing long reasoning chains in speech form and causing latency for complex queries. Our work internalizes such reasoning as \emph{inner thinking}, avoiding unnecessary speech synthesis while maintaining reasoning capabilities.

\textbf{Inference Scaling and Long-Chain Reasoning.}   Chain-of-Thought (\textsc{CoT}) prompting is a seminal technique for reasoning in large language models~\citep{cot}. By instructing the model to \emph{“please think step by step”}, CoT externalizes internal reasoning, breaking complex problems into smaller, tractable subproblems and significantly enhancing LLM performance. Building on this, OpenAI’s o1~\citep{o1} model introduced Test-time-scaling for advanced multi-step reasoning. DeepSeek-R1~\citep{deepseek-r1} leveraged a reinforcement learning approach, GRPO algorithm, inspiring subsequent methods such as \textsc{DAPO}~\citep{dapo} and \textsc{GSPO}~\citep{qwen3}. More recent work, including \textsc{Self-Evolving}~\citep{selfevolving} and \textsc{Latent Reasoning}~\citep{latentreasoning}, further improves reasoning flexibility and robustness. The latest efforts extend these reasoning techniques to multimodal settings~\citep{llava-cot,visionr1,audio-reasoner,r1-aqa,han2025videoespresso}.

\textbf{Reasoning Efficiency.}  Long reasoning chains introduce significant latency, posing a critical challenge for real-time applications. In text LLMs, efficiency has been improved through instruction fine-tuning with controlled output lengths, shorter chains of thought, and token- or draft-level compression strategies (e.g., \textsc{TokenSkip}~\citep{tokenskip}, \textsc{Chain-of-Draft}~\citep{chainofdraft}). Reinforcement learning methods, such as \textsc{Thinkless}~\citep{thinkless}, \textsc{ConciseRL}~\citep{conciserl}, and \textsc{LCPO}~\citep{L1}, further enable adaptive control over reasoning length, while hybrid models decide when to reason or respond directly, mitigating overthinking~\citep{thinkonluyouneed,longshortcot}. However, these approaches assume users can freely browse model outputs. In contrast, speech models generate outputs in a time-linear fashion, where long reasoning chains directly increase latency. This motivates \textsc{Mini-Omni-Reasoner}’s token-level \emph{“thinking-in-speaking”} paradigm, which internalizes reasoning without producing unnecessary speech.

\section{Conlusion}

In this work, we present \textsc{Mini-Omni-Reasoner}, a framework that simulates the human-like interplay between complex inner reasoning and outward verbalization through a novel Thinking-in-Speaking formulation. Unlike conventional methods that adopt a rigid thinking-before-speaking paradigm and suffer from high response latency, our approach interleaves unspoken reasoning tokens with spoken response tokens at the token level. This design produces a mixed sequence of reasoning and output, enabling the model to generate fluent and timely speech while preserving logical consistency and semantic coherence. To support this framework, we introduce the \textsc{Spoken-Math-Problems-3M} dataset, tailored to train and evaluate such interleaved reasoning–speech generation. Through careful modeling and alignment strategies, \textsc{Mini-Omni-Reasoner} adheres more closely to natural communication patterns, maintaining prosody while enhancing reasoning quality. Comprehensive evaluations on the Spoken-MQA benchmark show that our model achieves notable gains in both arithmetic and contextual reasoning, while reducing output length and eliminating decoding latency. These results demonstrate that fluent spoken interaction and high-quality reasoning can be jointly realized within a unified architecture, opening new directions for real-time, reasoning-aware speech systems. The code and data will be gradually open-sourced on our project homepage.

\section*{Acknowledgment}
% We 非常感谢xxx对论文的orgnization以及写作上提出的宝贵意见,为论文的构思提出了启发性的意见. 也感谢为Spoken-Math-Problems数据库提供数据源支撑的xxx，xxx,xxx 工作的贡献.
We sincerely thank \textbf{Changqiao Wu} for his valuable feedback on both the technical details and engineering implementation of this work. We also acknowledge the contributions of Orca-Math~\citep{orcamath}, MetaMath~\citep{metamath}, GSM8K~\citep{gsm8k}, and SimpleOP~\citep{singleop}, which provided critical data sources for the construction of the \textsc{Spoken-Math-Problems-3M} dataset.

\bibliography{iclr2025_conference}
\bibliographystyle{iclr2025_conference}

\end{document}